\documentclass[10pt,twocolumn,letterpaper]{article}

\usepackage{iccv}
\iccvfinalcopy

\usepackage{times}
\usepackage{epsfig}
\usepackage{graphicx}
\usepackage{amsmath}
\usepackage{amssymb}
\usepackage{balance}

\usepackage[font=footnotesize,labelfont=bf]{caption}

\usepackage{diagbox}
\usepackage{float}
\usepackage{stfloats}

\usepackage{subfigure}

\usepackage{enumerate}
\usepackage{multirow}
\usepackage{tabularx}
\usepackage{dsfont}
\usepackage{url}

\usepackage{color}

\usepackage{amsthm}
\usepackage[ruled,vlined]{algorithm2e}
\usepackage[export]{adjustbox}
\usepackage{comment}

\usepackage[utf8]{inputenc} 
\usepackage[T1]{fontenc}    
\usepackage{url}            
\usepackage{booktabs}       
\usepackage{amsfonts}       
\usepackage{nicefrac}       
\usepackage{microtype}      

\usepackage{accents}

\usepackage{enumitem}
\usepackage{bm}

\usepackage[numbers]{natbib}
\usepackage{makecell}
\renewcommand{\paragraph}[1]{\vspace{1.25mm}\noindent\textbf{#1}}

\definecolor{dg}{rgb}{0,0.694,0.298}
\definecolor{purple}{rgb}{0.4,0.176,0.569}

\usepackage{pifont}
\newcommand{\cmark}{\textcolor{dg}{\ding{52}}}%
\newcommand{\xmark}{\textcolor{red}{\ding{56}}}%

\usepackage[pagebackref=true,breaklinks=true,colorlinks,bookmarks=false]{hyperref}

\def\iccvPaperID{9569} 

\ificcvfinal\fi

\begin{document}

\title{What can Discriminator do? Towards Box-free Ownership Verification of Generative Adversarial Networks}

\author{Ziheng Huang$^{1\dagger}$, Boheng Li$^{1\dagger}$, Yan Cai$^{1}$, Run Wang$^{1*}$, Shangwei Guo$^2$, \\ Liming Fang$^{3}$, Jing Chen$^{1}$, Lina Wang$^{1}$ \\
        $^1$ Key Laboratory of Aerospace Information Security and Trusted Computing,\\ Ministry of Education, School of Cyber Science and Engineering, Wuhan University, China \\
	$^2$ College of Computer Science, Chongqing University, China \\
        $^3$ College of Computer Science and Technology, Nanjing University of \\ Aeronautics and Astronautics, China \\
        \small $^\dagger$ Equal contribution $^*$ Corresponding author. E-mail: \tt wangrun@whu.edu.cn}

\maketitle

\begin{abstract}
In recent decades, Generative Adversarial Network (GAN) and its variants have achieved unprecedented success in image synthesis. However, well-trained GANs are under the threat of illegal steal or leakage. The prior studies on remote ownership verification assume a  black-box setting where the defender can query the suspicious model with specific inputs, which we identify is not enough for generation tasks. To this end, in this paper, we propose a novel IP protection scheme for GANs where ownership verification can be done by checking outputs only, without choosing the inputs (\ie{}, box-free setting). Specifically, we make use of the unexploited potential of the discriminator to learn a hypersphere that captures the unique distribution learned by the paired generator. Extensive evaluations on two popular GAN tasks and more than 10 GAN architectures demonstrate our proposed scheme to effectively verify the ownership. Our proposed scheme shown to be immune to popular input-based removal attacks and robust against other existing attacks. The source code and models are available at \href{https://github.com/AbstractTeen/gan\_ownership\_verification/}{\textcolor{magenta}{https://github.com/AbstractTeen/gan\_ownership\_verification.}} 

\end{abstract}


\section{Introduction}\label{sec:intro}

With the rapid development of GANs, we have witnessed fruitful applications of GAN in many fields, such as realistic facial images synthesis \cite{skorokhodov2022stylegan}, fine-grained attribute editing \cite{wang2022high}, \etc. Unlike the classification model with specified label prediction, the GANs learn a data distribution and output the synthesized data sample within a certain distribution. In GANs, the discriminator and generator are two essential components, where the discriminator works as a judger to discriminate whether the sample is produced by the generator, the generator learns to generate more realistic samples to confuse the discriminator \cite{goodfellow2020generative,cai2021generative}. Usually, the discriminator is discarded after training since the generator is the core asset for synthesizing high-quality images.

\begin{figure}
    \centering
    \includegraphics[width=0.9\linewidth]{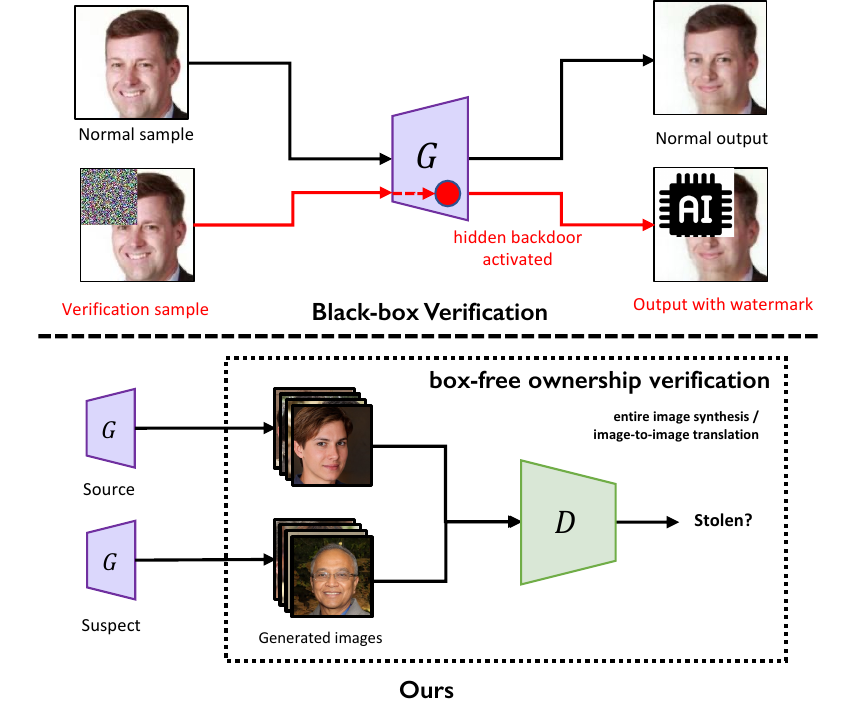}
    \vspace{-15pt}
    \caption{Comparison of the verification process between previous black-box watermark-based verification paradigm~\cite{ong2021protecting} and our box-free method. In the black-box setting, carefully-crafted verification samples should be fed to the suspicious model to activate a hidden backdoor in the model (the red circle) and generate watermarked outputs. However, in box-free setting, querying the model with deterministic inputs is not allowed. Ownership verification should be done with only output images.}
    \label{fig:teaser}
    \vspace{-15pt}
\end{figure}

Training a decent GAN requires a huge investment of resources, such as computing resources, labeled/unlabeled training dataset, time, and human labors \cite{lin2021anycost}. However, well-trained generators are under the threat of unintentional leakage and theft. The adversary may deploy the stolen model on the Internet for profit and the owner (also the \textit{defender}) is only able to verify the ownership remotely by querying the suspicious model \cite{lukas2022sok,ong2021protecting,juefei2022countering,gan2023towards}.

Most existing works on IP protection of DNN models assume the owner can query the suspicious model with specific inputs (\ie{}, black-box setting). Based on this fundamental assumption, two schools of solutions have been proposed: model watermark \cite{zhang2018protecting,ong2021protecting,li2022untargeted} and model fingerprint \cite{peng2022fingerprinting,lukas2019deep}. They either proactively embed or passively extract a hidden functionality in the model, where outlier outputs can be activated by a specific query set (known as the \textit{verification set}). For verification, the defender queries the model with this set and observes whether the outputs match the source model. Since it is improbable for any other model to perform the same abnormal behavior, the owner can judge whether the suspicious model is a stolen copy.

However, the black-box assumption is challenged in generation tasks. For example, the whole-image synthesis task takes a randomized latent representation as input. In reality, this representation is usually sampled from a pre-defined distribution, such as normal distribution. Therefore, the adversaries can prohibit the verification by sampling the latent representation themselves. The black-box methods are also shown to be vulnerable to input transformation-based removal attacks \cite{guo2021fine,wang2022rethinking}. Moreover, recent literature has shown the outlier verification samples can be detected, inspected, or reverse-engineered \cite{guo2023scale,wang2022rethinkingtrojan}. Note that the verification set plays a similar role to private keys in cryptography \cite{jia2021entangled,li2023black}. Once it is disclosed, the adversaries are capable to launch ambiguity attacks, or invalidate the watermark/fingerprint via methods such as adversarial training. These limitations inspire us to raise an important question: \textit{can ownership verification be done via checking outputs only, without choosing the inputs (\ie{}, box-free setting)?} This setting is more challenging because queries made by the defender are totally equivalent to those of normal users. There is no chance to activate a hidden functionality. 
    
In this paper, we make the first attempt to box-free ownership verification of GANs. Based on the fact that GAN suffers from unstable training, we reveal the unexploited potential of the discriminator to capture the model-specific distribution learned by the paired generator. We utilize the discriminator's representations to learn a network featuring a hypersphere that encloses the distribution learned by the generator. Our proposed scheme does not require specifying any input nor training additional detection networks, the ownership verification can be done effectively via feeding a batch of suspicious images to the learned network. However, due to the gradual degradation issue, it is challenging for the discriminator to extract meaningful feature representations without sacrificing the performance of the generator. To tackle this problem, we leverage the pearson correlation coefficient \cite{pearson1896vii} to quantify the implicit reconstruction ability of the discriminator, and prevent the degradation via adding the term into the loss function of the discriminator.

To comprehensively evaluate the effectiveness and robustness of our proposed method, we conduct experiments on two popular GAN tasks (\ie{}, entire image synthesis, image-to-image translation) and 10 state-of-the-art GAN architectures to demonstrate the effectiveness of our scheme in verifying the ownership of generator. We also show qualitatively and quantitatively that our scheme is immune to popular and powerful removal attacks (\eg{}, input transformation-based and reverse engineering-based attacks) and robust to other existing attacks.

Our main contributions are summarized as follows:

\begin{itemize}[leftmargin=*]
\item We identify a fundamental limitation of black-box setting-based ownership verification schemes on generation tasks, \ie{}, choosing deterministic inputs is not allowed for applications like unconditioned image synthesis.

\item We reveal the unexploited potential of the well-trained discriminator for capturing the unique distribution learned by the paired generator. Based on this finding, we make the first attempt towards box-free verification scheme of GANs, which does not require specifying the input and does not rely on additional models.

\item Extensive evaluations on two popular GAN applications and more than 10 GAN architectures demonstrate our proposed scheme to effectively verify the ownership. Through qualitative and quantitative analysis, we show that our proposed scheme is immune to popular removal attacks and robust to other existing attacks.
\end{itemize}

\section{Related Work}\label{sec:related}

\subsection{Ownership Verification} \label{sec:ip_pro}

\textbf{Black-box Ownership Verification.} The black-box ownership verification scheme assumes the defender can verify the ownership via querying the suspicious model with specific inputs \cite{adi2018turning,zhang2018protecting,le2020adversarial,wang2023free}. Towards this end, the owner either predetermines and trains the model to learn a set of abnormal input-output behaviours (model watermarks), or extract a set of boundary samples (\eg{}, adversarial examples) that can identify the model (model fingerprint). Then, this hidden behavior can be extracted remotely by querying the model with these special inputs. Recently, \citet{ong2021protecting} proposed the first work towards ownership verification of GANs. In their work, the IP information was embedded through backdooring the generator, after which trigger inputs (\ie{}, carefully-crafted inputs with trigger noises, as shown in Fig. \ref{fig:teaser} top panel) will result in a visible watermark like a company's logo on generated outputs. However, for some GAN applications (\eg{}, unconditional image synthesis), black-box is not enough since deterministic inputs are not allowed \cite{yu2021artificial}. Moreover, the recent removal attacks have shown great threats to the survival of backdoor or adversarial input-based black-box verification paradigms.

\textbf{Removal Attacks}. The black-box methods rely on hidden functionalities in the model. Therefore, current attacks aim to remove or avoid activating this hidden functionality. A straightforward attack is to eliminate the hidden functionality through model modifications like pruning \cite{lukas2022sok}. A more targeted attack is first to reverse-engineer the verification samples and then invalidate the watermarks/fingerprints through adversarial training \cite{wang2022rethinkingtrojan,tao2022better}. Observing the verification samples are less robust than normal inputs, another attack is to evade verification via preprocessing the input \cite{guo2021fine,wang2022rethinkingtrojan}. This attack has become the recent trend since it is model architecture-careless and not limited to a specific watermarking technique. It is also intractable as the input transformations can be diverse and usually hard to consider in advance. In contrast, our proposed box-free ownership verification scheme is free from choosing verification samples thus totally immune to the intractable input transformation-based attacks and reverse engineering-based attacks.

\begin{table*}[t]
\scriptsize
\centering
\caption{Comparison of our approach with prior works. In the ``Img. Syn./Trans.'' column, \xmark~ denotes not applicable or not evaluated in the original paper.}
\vspace{-5pt}
\setlength{\tabcolsep}{2.8pt}
\begin{tabular}{c||c|c|c|c|c|c|c|c}
\toprule
Method & Year & Technique & \makecell{Target \\ model} & Purpose & Box-free? &  \makecell{Img. Syn./ \\ Trans.?} & \makecell{Ambiguity \\ attack?} & \makecell{No external \\ model?}  \\ \midrule
\citet{uchida2017embedding} & 2017 & Model Watermarking & DNN & Ownership Verification & \xmark  & --- & \xmark & \cmark  \\
\citet{adi2018turning} & 2018 & Model Watermarking & Classifiers & Ownership Verification & \xmark  & --- & \xmark & \cmark  \\
\citet{zhang2018protecting} & 2018 & Model Watermarking & Classifiers & Ownership Verification & \xmark  & --- & \xmark & \cmark  \\
\citet{rouhani2019deepsigns} & 2019 & Model Watermarking & DNN & Ownership Verification & \xmark &  --- & \xmark & \cmark \\
\citet{lukas2019deep} & 2019 & Model Fingerprinting & Classifiers & Ownership Verification & \xmark &  --- & \xmark & \cmark \\
\citet{le2020adversarial} & 2020 & Model Watermarking & Classifier & Ownership Verification & \xmark &  --- & \xmark & \cmark \\
\citet{zhao2020afa} & 2020 & Model Fingerprinting & Classifiers & Ownership Verification & \xmark &  --- & \xmark & \cmark \\
\citet{jia2021entangled} & 2021 & Model Watermarking & Classifier & Ownership Verification & \xmark &  --- & \cmark & \cmark \\
\citet{cao2021ipguard} & 2021 & Model Fingerprinting & Classifiers & Ownership Verification & \xmark &  --- & \xmark & \cmark \\
\citet{bansal2022certified} & 2022 & Model Watermarking & DNN & Ownership Verification & \xmark &  --- & \cmark & \cmark  \\

\citet{peng2022fingerprinting} & 2022 & Model Fingerprinting  & Classifiers & Ownership Verification & \xmark & --- & \xmark & \cmark  \\
\citet{yangmetafinger} & 2022 & Model Fingerprinting  & Classifiers & Ownership Verification & \xmark & --- & \xmark & \xmark  \\
\citet{yu2019attributing} & 2019 & Fingerprint Extraction & GAN & Model Attribution & \cmark &  \cmark~/~\xmark & \xmark & \xmark \\

\citet{yu2021artificial} & 2021 & GAN Fingerprinting  & GAN & Model Attribution & \cmark & \cmark~/~\cmark & \cmark & \xmark \\ 
\citet{Girish_2021_ICCV} & 2021 & Fingerprint Extraction & GAN & Model Attribution & \cmark & \cmark~/~\cmark & \xmark & \xmark \\ 
\citet{asnani2021reverse} & 2021 & Fingerprint Extraction  & GAN & Model Attribution & \cmark & \cmark~/~\cmark & \xmark & \xmark  \\ 
\citet{yu2022responsible} & 2022 & GAN Fingerprinting  & GAN & Model Attribution & \cmark &\cmark~/~\xmark & \cmark & \xmark \\
\citet{guarnera2022exploitation} & 2022 & Fingerprint Extraction  & GAN & Model Attribution & \cmark & \cmark~/~\xmark & \xmark & \xmark \\ 
\citet{ong2021protecting} & 2021 & GAN Watermarking  & GAN & Ownership Verification & \xmark & \cmark~/~\cmark & \cmark & \cmark \\
\midrule

Ours & 2023 & Distribution Capturing & GAN & Ownership Verification & \cmark & \cmark~/~\cmark  & \cmark & \cmark \\

\bottomrule
\end{tabular}
\label{Table:overview}
\vspace{-10pt}
\end{table*}

\textbf{Ambiguity Attack}. Recent works \cite{fan2019rethinking,fan2022deepipr,ong2021protecting} revealed the concept of ambiguity attack, where it is proved that unless an irreversible verification scheme is adopted, the adversary can forge his/her own vouch using exactly the same technique the owner adopted. Subsequently, when one claims ownership of the model, the adversary can also claim the ownership due to the existence of his/her own vouch. Finally, the ownership is in doubt. To ensure the owner is free from this concern, a feasible technique used for verifying ownership should necessarily be non-reproducible, even if the adversary has full control of the stolen model and acquires knowledge of the adopted ownership verification paradigm. A practical and robust ownership verification scheme should well survive these two threats.

\subsection{Model Attribution}
The model attribution was initially developed to combat DeepFakes, where researchers focus on attributing certain fake images to the specific types of GAN that generated them \cite{yu2019attributing,yu2021artificial,yu2022responsible,albright2019source,guarnera2022exploitation}. 
Generally, the attribution techniques aim to analyse the unique fingerprints carried by the GAN-generated images, or proactively watermark the output images through methods like steganography \cite{wu2020watermarking,zhang2020model,zhang2021deep}. Due to the merit that these techniques can usually work in the box-free setting where only generated images are available, it has shown potential or even applications in ownership verification of GANs. However, the existing studies on GAN model attribution are not ready for this challenging task, since i) most works are limited to classifying the model architecture and/or datasets~\cite{bui2022repmix,Girish_2021_ICCV,yang2022deepfake} rather than a specific GAN model; ii) nearly all attribution techniques require training a powerful external classifier, which is both time and resource consuming; and iii) most works' external classifiers can be trained with only real/fake images~\cite{yu2019attributing,asnani2021reverse,ding2021does}, and the steganography-based techniques could be easily reproduced by the attacker \cite{wu2020watermarking,zhang2020model}, which denotes that the adversary can easily forge the verification vouch and perform ambiguity attacks or overwriting attacks. A comparison between our approach with prior studies is illustrated in Tab.~\ref{Table:overview}.

\section{The Proposed Verification Scheme}\label{sec:method}
Motivated by the unstable training phenomenon of GAN training, the insight behind our proposal is to capture the model-specific distribution by exploiting the potential of the paired discriminator. Before stepping into the details of our proposed scheme, we first introduce the organization of this section. First, we formalize our threat model. Then, we describe the details of the essential loss terms used in our proposed scheme. Finally, we introduce the pipeline and training flow of our proposed approach.

\subsection{Threat Model}
In our threat model, two opposing parties are considered. A \textit{model owner} (also the \textit{defender}) who trains the source model, and an \textit{adversary} who has stolen the source model through illegal means and deploys the piracy model on Internet APIs publicly accessible for profit.

\paragraph{Defender's Goals and Capabilities.} The goal of the defender is to identify the stolen models that are remotely deployed by the attacker. The defender (1) has white-box access to the source model, (2) can query the suspicious model but, (3) can not specify the query input (\ie{}, box-free setting). This setting is more challenging than the black-box setting widely adopted by prior works \cite{adi2018turning,ong2021protecting,wang2023free}, as specifying carefully-crafted inputs is not allowed.

\paragraph{Adversary's Goals and Capabilities.} The adversary's goal is not to be verified as pirated while keeping the piracy a similar performance as the source model. For this purpose, the adversary  may modify the model (model modification) or manipulate the inputs/outputs (sample transformation). Observe that our scheme does not suffer from the intractable input transformation-based attacks since our inputs are totally equivalent to those of normal users. Adopting such attacks would only impair the performance yet does not bring any benefits. The assumptions we make are standard and are also widely adopted by prior works \cite{adi2018turning,zhang2018protecting,jia2021entangled,ong2021protecting}.

\subsection{Compactness Loss}\label{sec:comloss}
The discriminator witnessed the gradual development of the generator and has comparable parameters to the latter. It learns a hyperplane in its embedding space to distinguish between real and generated images. Intuitively, it potentially learned how to extract special representations in images generated by the paired generator. However, in real-world scenarios, suspicious images are usually generated by unknown and unseen GANs. Since embedding spaces of different generators are not aligned, it is difficult to harness the discriminator and find maximum margin hyperplanes to distinguish between different GAN instances.

Inspired by previous works on data description \cite{tax1999support,tax2004support}, we propose to separate the data via optimizing a hypersphere instead of a hyperplane. Specifically, let $\mathcal{X} \subseteq \mathbb{R}^{c\times h\times w}$ be the data space, and $\phi: \mathbb{R}^{c \times h \times w} \rightarrow \mathbb{R}^d$ be the ``encoder'' part of the discriminator (explained later), which maps the data to a $d$-dimensional feature space. Our objective is to find the smallest hypersphere specified by a center $\boldsymbol{c}\in\mathbb{R}^d$  and radius $R > 0$ that encloses the majority of the data distribution of the paired generator in feature space. Therefore, our main objective is to minimize the ``compactness'' of representations \cite{perera2019learning,ruff2018deep,reiss2021panda}. Given the data $\{\boldsymbol{x}_1, \dots, \boldsymbol{x}_n\}$ from the paired generator, our objective is defined as:
\begin{equation}
\setlength\abovedisplayskip{3pt}
\setlength\belowdisplayskip{0.5pt}
    \label{eqn:cl}
    \min_{\mathcal{W}, R} \quad  R^2 + \frac{1}{\nu n} \sum_{i=1}^{n}\max\{0, \left \| \phi(\boldsymbol{x}_i;\mathcal{W}) -\boldsymbol{c} \right \|^2 - R^2\}
\end{equation}
where $\mathcal{W}$ indicates the weights of $\phi$.  This objective contains two terms. The first term aims to minimize the radius thus volume of the hypersphere, while the second term penalizes the points outside the hypersphere. $\nu \in (0,1]$ is a hyperparameter that balances the importance of the two terms. We empirically set $\nu$ to $0.35$ in the following experiments.

The benefits of our approach are as follows. First, our training is unsupervised, \ie{}, only requires data generated by the paired GAN, analogous to the training of GAN itself and does not introduce additional annotation overhead. Second, prior works utilizing the compactness loss as their objective are shown vulnerable to “hypersphere collapse” where the hypersphere radius collapses to zero and the network converges to trivial solutions \cite{tian2022tvt}. Fixing $\boldsymbol{c}$ as the mean of the network representations is shown to be helpful to avoid overfitting and hypersphere collapse \cite{ruff2018deep,chong2020simple}. However, it is still difficult to define a proper hypersphere center $\boldsymbol{c}$ with an initial network whose parameters are random. In contrast, our well-trained discriminator network potentially provides us with a robust network representation thus a robust $\boldsymbol{c}$. Empirical results show that this strategy makes the convergence faster and more robust, the hypersphere collapse hardly exist.

\subsection{Pearson Correlation Loss} \label{method:ploss}
Recall that we expect our discriminator to provide a robust initial center for training. This is feasible since we assume the discriminator extracts data representations well, thus network representations are useful. However, in reality, we observe that the discriminator usually converges to a constant and no helpful network representation is preserved. This is not surprising since the more the GAN is trained, the less possible to distinguish between samples of real and generated data. The optimal discriminator converges to $1/2$ and thus no extraction ability is preserved, as many existing literature has pointed out \cite{goodfellow2020generative,zhang2018convergence}.

To tackle this challenge, we propose to preserve the useful network representations of the discriminator via encouraging it to implicitly reconstruct the ground truth latent representation ${\boldsymbol{z}}$. Our key insight is that this additional task encourages the discriminator to fit the generator's latent distribution and prohibits its trivial convergences. However, we empirically found that the straightforward MSE loss makes the training extremely unstable. In our approach, we leverage the pearson correlation loss, which is inspired by the pearson Correlation Coefficient (PCC) \cite{pearson1896vii}, to measure the quality of the reconstructed latent representation. We define our term as:
\begin{equation}
\setlength\abovedisplayskip{3pt}
\setlength\belowdisplayskip{0.5pt}
\rho(\boldsymbol{z}, \hat{\boldsymbol{z}}) = \frac{\sum_{i=1}^{n_z}(\boldsymbol{z}_i - \mu_z)(\hat{\boldsymbol{z}_i}- \mu_{\hat{z}})/n_z}{\sigma_z \sigma_{\hat{z}}}
\end{equation}
where $\hat{\boldsymbol{z}}$ is the reconstructed latent representation and ${\boldsymbol{z}}$ is the ground truth. ${n_z}$ is the dimension of $z$. $ \mu$ and $\sigma$ indicates the mean value and standard deviation, respectively. $\rho(\cdot, \cdot)\in [-1, 1]$ measures the linear correlation between the variables. The higher $\rho$ indicates better reconstruction performance.

PCC is “milder” than MSE \cite{su2019gan}. The pearson correlation loss ensures that the discriminator can implicitly reconstruct the latent representation from its corresponding generated image while avoiding making the training unstable. For training, obviously, this additional task can be easily cooperated with the BCE loss of the native GAN training. There is no annotation cost and no notable training overhead. We show in the \textcolor{magenta}{supplementary materials} that this additional loss does not bring any degradation to the original generation task.

Note that this loss is optional for GANs that are trained in supervised setting (\eg{}, StarGAN). This is because these tasks usually require the discriminator to do an additional classification task, therefore the aforementioned convergence problem does not necessarily exist.

\subsection{Training Pipeline}
The whole training pipeline of our scheme is presented in Algorithm \ref{alg:pipeline}. We utilize the pearson correlation loss to preserve the network representations of the well-trained discriminator and harness these representations to train a robust hypersphere in the embedding space which captures the unique data distribution of the paired generator. We show the whole training pipeline as follows.

\paragraph{Step 1. Redefine the Training Objective:} For unsupervised GANs, we introduce an additional task (\ie{}, reconstructing the latent representation) to ensure the network representations of the well-trained discriminator are useful. The native GAN is composed of a generator mapping $\mathbb{R}^{d} \rightarrow \mathbb{R}^{c\times h\times w}$ and a discriminator mapping $\mathbb{R}^{c\times h\times w} \rightarrow \mathbb{R}$, which can be divided into an ``encoder'' $\phi: \mathbb{R}^{c\times h\times w} 
\rightarrow \mathbb{R}^d$ and a ``classifier'' that maps $\mathbb{R}^{d} 
\rightarrow \mathbb{R}$. As explained in Sec. \ref{method:ploss}, we wish the reconstructed latent representation ${\hat{\boldsymbol{z}}} = \phi(G(\boldsymbol{z});\mathcal{W}) \in \mathbb{R}^d$ to be close to the ground truth representation ${{\boldsymbol{z}}} \in \mathbb{R}^d$. This is done via adding the pearson correlation loss to the adversarial objective of both $G$ and $D$. The training objective of the generator is:
\begin{equation}
\setlength\abovedisplayskip{3pt}
\setlength\belowdisplayskip{0.5pt}
    \min_{G} \mathcal{L}_{G} = \mathbb{E}_{\bm{z} \sim p_{\bm{z}}(\bm{z})}[\log (1 - D(G(\bm{z})))- \lambda\rho(\boldsymbol{z}, \hat{\boldsymbol{z}})]
\end{equation}
The training objective of the discriminator is:
\begin{equation}
\setlength\abovedisplayskip{3pt}
\setlength\belowdisplayskip{0.5pt}
    \max_{D} \quad \mathcal{L}_{D} = \mathbb{E}_{\bm{x} \sim p_{\mathcal{D}}(\bm{x})}[\log D(\bm{x})- \lambda\rho(\boldsymbol{z}, \hat{\boldsymbol{z}})] 
\end{equation}
where $\lambda$ is a hyper-parameter that adjusts the pearson correlation strength. We empirically set $\lambda=0.5$ in experiments.

\paragraph{Step 2. Train the GAN Models:} We then initialize and train the GAN models $G$ and $D$ with some training data  $\mathcal{D} = \{\boldsymbol{x}_1, \dots, \boldsymbol{x}_n\} \subseteq \mathcal{X}$ until we reach the maximum epochs. Except for the additional pearson correlation loss, the training of the GANs follow the settings in their original papers. After this, the generator is capable to tackle the task distribution and can be deployed. 

\paragraph{Step 3. Empower the Discriminator:} The final step is to utilize the discriminator's well-trained network representations and optimizes the network which forms a hypersphere to enclose the learned distribution. We first harness the generator to generate a few data points sampled from the distribution, forming a set of training data $\mathcal{D}^\prime = \{\boldsymbol{x}^\prime_1, \dots, \boldsymbol{x}^\prime_n\} \subseteq \mathcal{X}$, where $\boldsymbol{x}^\prime_i = G(\boldsymbol{z})$, $\boldsymbol{z} \sim p_{\bm{z}}(\bm{z})$. As explained in Sec. \ref{sec:comloss}, we fix $\boldsymbol{c}$ as the mean of the network representations of the discriminator and train the network with data $\mathcal{D}^\prime$ and the objective described in Eq. (\ref{eqn:cl}).

\vspace{10pt}
We use stochastic gradient descent (SGD) to optimize the parameters $\mathcal{W}$ of the neural network with backpropagation. Noticeably, using one common SGD learning rate may be inefficient to optimize $\mathcal{W}$ and ${R}$ simultaneously since they usually have different scales, as \citet{ruff2018deep} pointed out. Therefore, we optimize $\mathcal{W}$ and ${R}$ alternatively as suggested. In detail, we first fix the radius $R$ and train the network parameters $\mathcal{W}$ for every interval $k \in \mathbb{N}$ epochs. Then, after every $k$ epoch, we solve for radius $R$ via line search with the current network parameters. We train the network parameters $\mathcal{W}$ and radius $R$ until convergence.

\begin{algorithm}[t]
	\footnotesize
	\SetAlgoLined
	\SetKwInOut{Input}{Input}
	\SetKwInOut{Output}{Output}
	\Input{Training data $\mathcal{D}$, Iteration $K$, Interval $k$, Learning rates $\tau$ and $\tau^\prime$.}
	\Output{Generator $G$, Network $\phi$ with parameters $\mathcal{W}$, center $\boldsymbol{c}$, and radius $R$.}
	
    Initialize the generator $G$ and discriminator $D$.\\
    \For{$i \in \{1...K\}$}{

        Get a batch of $\mathcal{D}$
        
        Calculate adversarial loss of $D$
        
        Update $D \leftarrow D - \tau\cdot\nabla_{D} \mathcal{L}_{D}$\\

        Calculate adversarial loss of $G$
       
        Update $G \leftarrow G - \tau\cdot\nabla_{G} \mathcal{L}_{G}$\\

	}

    \# Form a dataset $\mathcal{D}^\prime$ that consists of $n$  samples generated by $G$ \\
    $\mathcal{D}^\prime=\cup_{i=1}^nG(\boldsymbol{z})$ \\
    Initialize center $\boldsymbol{c}$, and radius $R$ \\
    step = 0 \\
    \While{loss not converge}{
        step += 1 \\
        Get a batch of $\mathcal{D}^\prime$ \\
        Calculate compactness loss:
        $$\quad \mathcal{L}_c =  R^2 + \frac{1}{\nu n} \sum_{i=1}^{n}\max\{0, \left \| \phi(\boldsymbol{x}_i;\mathcal{W}) -\boldsymbol{c} \right \|^2 - R^2\}$$
        Update $\mathcal{W} \leftarrow \mathcal{W} - \tau^\prime\cdot\nabla_{\phi} \mathcal{L}_c$\\
        \If{step \% $k$ == 0} {
        Update $R$ via line search
	}
    }
    \caption{Training Pipeline}
    \label{alg:pipeline}
\end{algorithm}
\vspace{-5pt}

\subsection{Ownership Verification}
In predicting a given input, we can calculate a score which measures the representation proximity to the captured unique distribution using the network parameters. Given an input $\boldsymbol{x}$, the representation proximity score is calculated by the distance of the point to the center of the hypersphere:

\begin{equation}
\setlength\abovedisplayskip{3pt}
\setlength\belowdisplayskip{0.5pt}
s(\boldsymbol{x}) = \left \| \phi(\boldsymbol{x};\mathcal{W}) -\boldsymbol{c} \right \|^2 - R^2
\end{equation}

The prediction is time and memory-efficient, since $s(\boldsymbol{x})$ is totally characterized by the network parameters $\mathcal{W}$, radius $R$, and the representation center $\boldsymbol{c}$. We do not require storing any other data for prediction and the prediction is done within a single forward pass. 

Note that $s(\boldsymbol{x})$ has different scales in different cases. Therefore, we feed a batch of images produced by the suspicious GAN and use Area Under Curve (AUC) score to measure the performance. This avoids selecting a deterministic proximity score threshold. Through extreme results on AUC scores (see Sec. \ref{sec:exp2}), we set the suspicious AUC score to $60\%$. That is, if a batch of suspicious images (batch size is empirically set to 500 in the following experiments) has an AUC $<60\%$, we claim ownership of the suspicious model.

\begin{figure}
     \centering
     \subfigure[AUC scores on LSUN dataset. ]{
         \centering
         \includegraphics[width=0.45\linewidth]{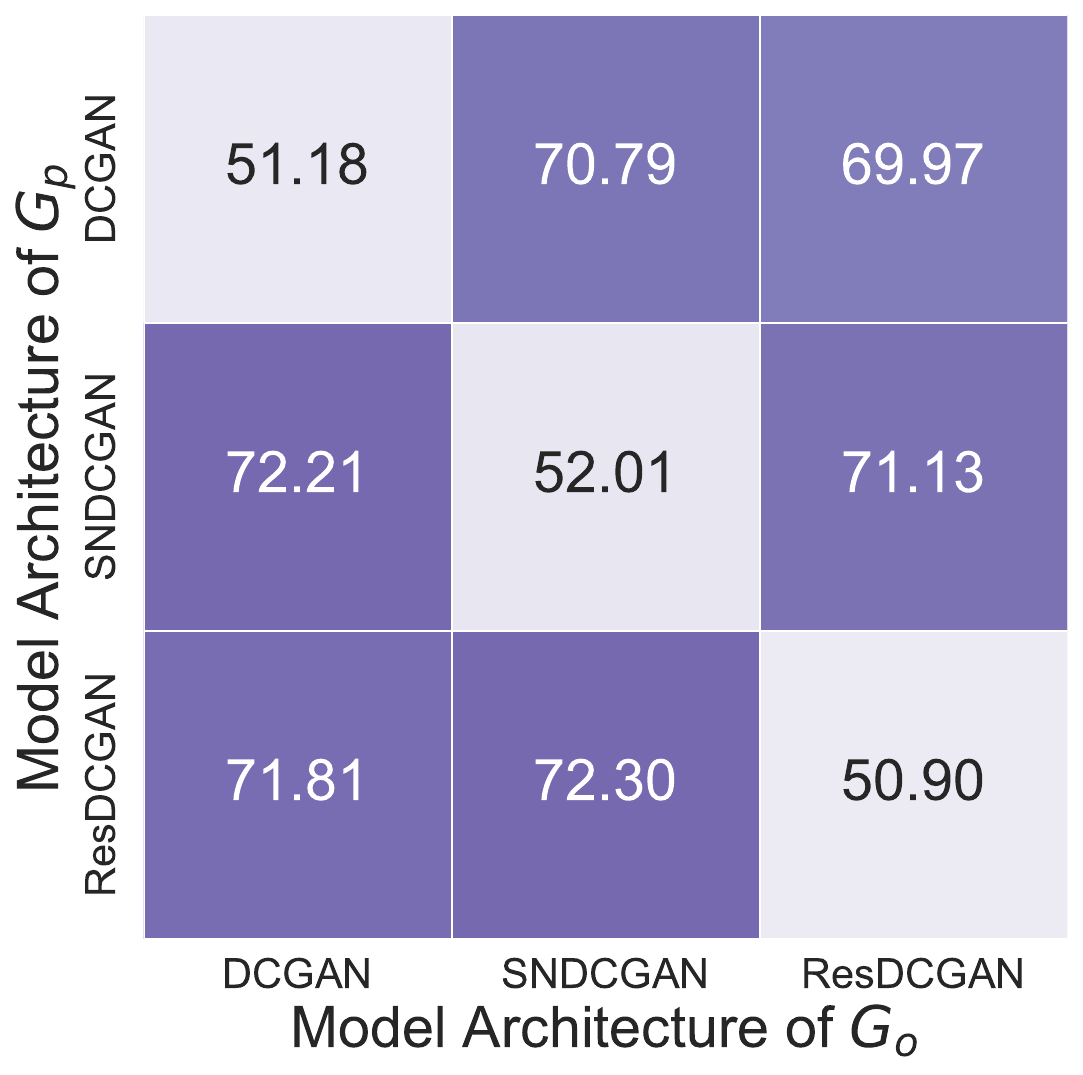}
         \includegraphics[width=0.45\linewidth]{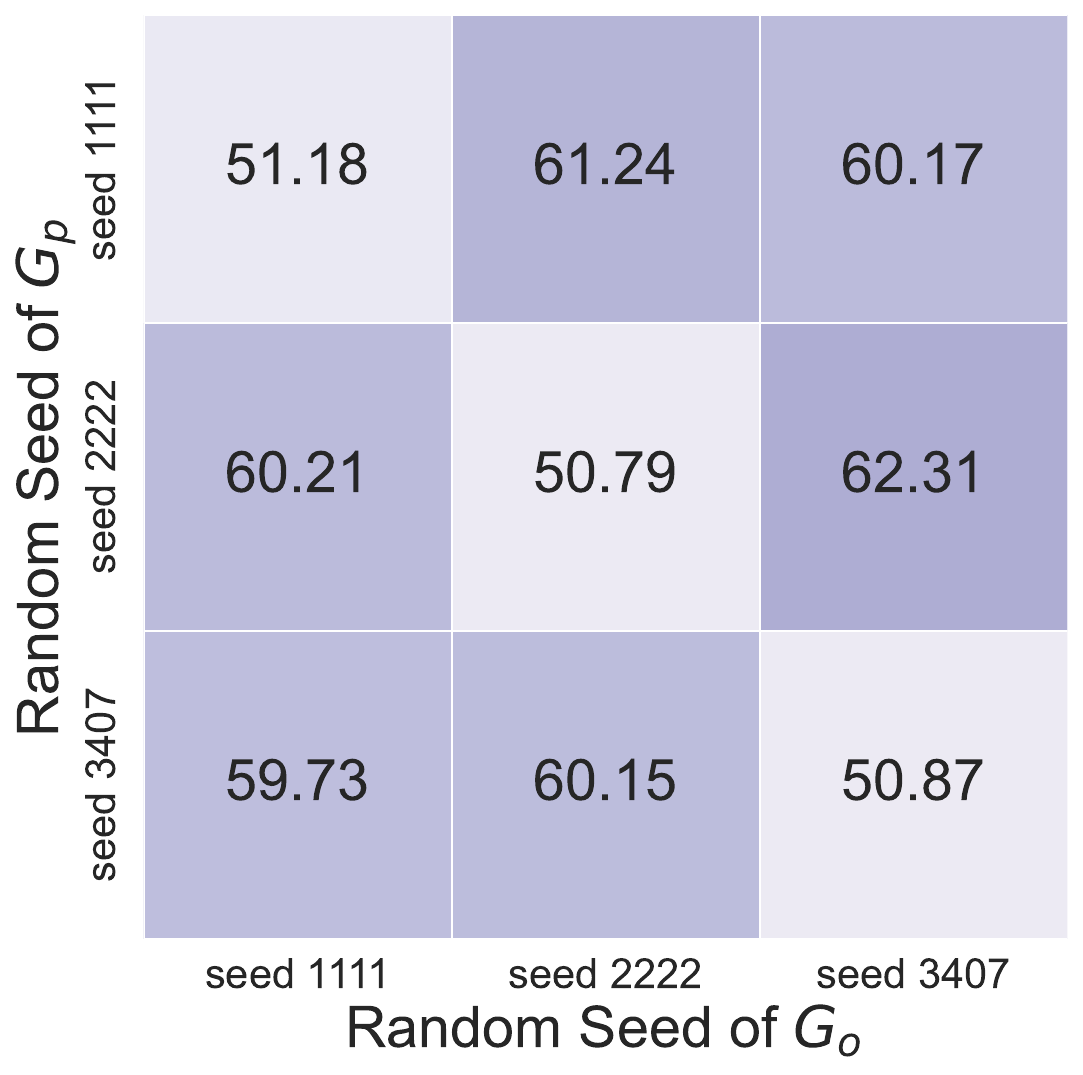}
         \includegraphics[height=0.45\linewidth]{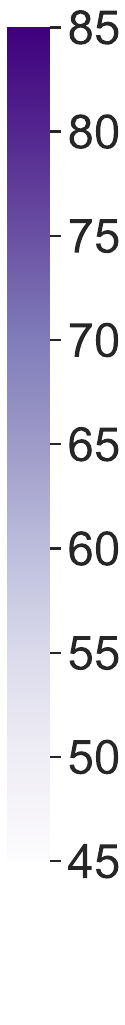}
         \vspace{-5pt}
         }
     \subfigure[AUC scores on CelebA dataset.]{
         \centering
         \includegraphics[width=0.45\linewidth]{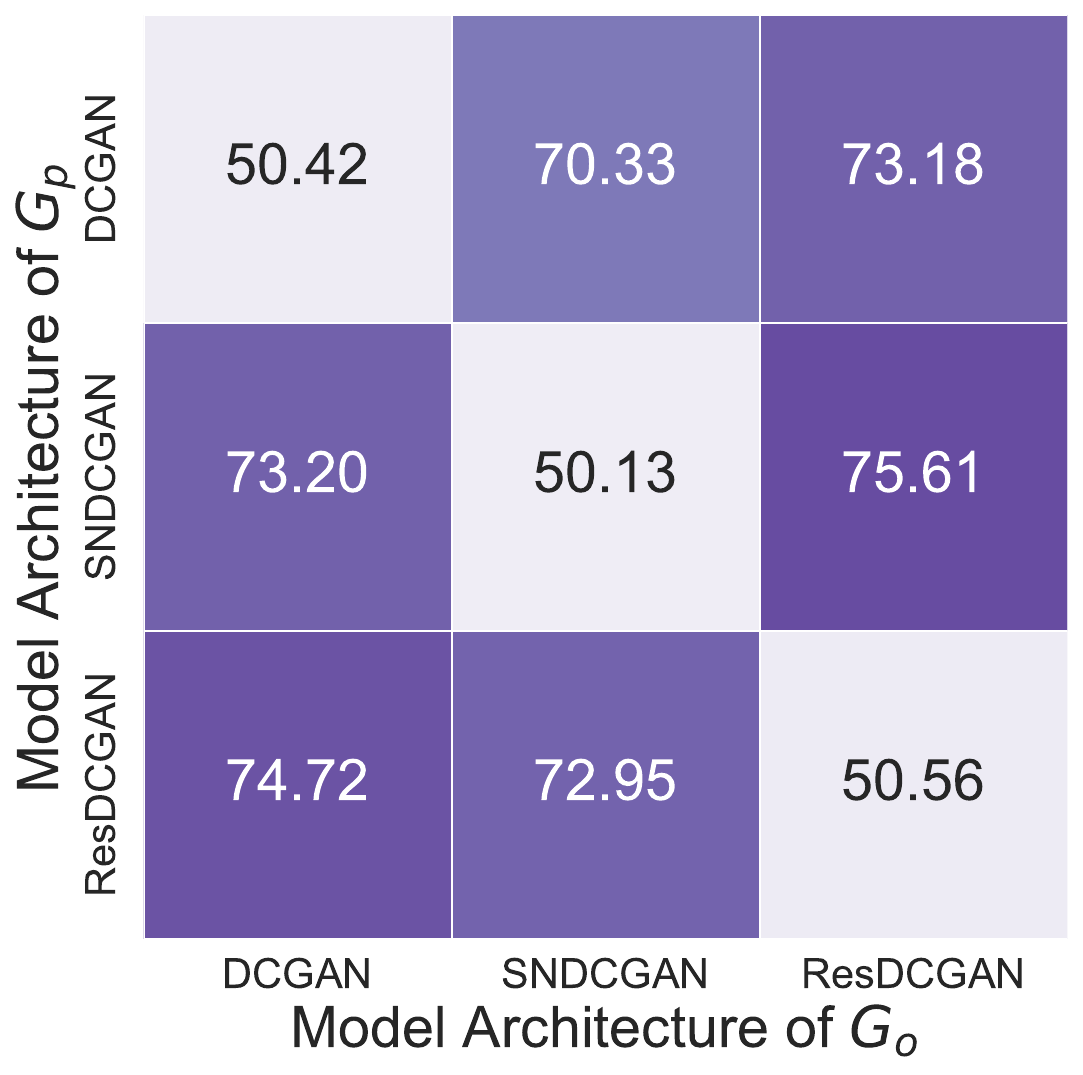}
         \includegraphics[width=0.45\linewidth]{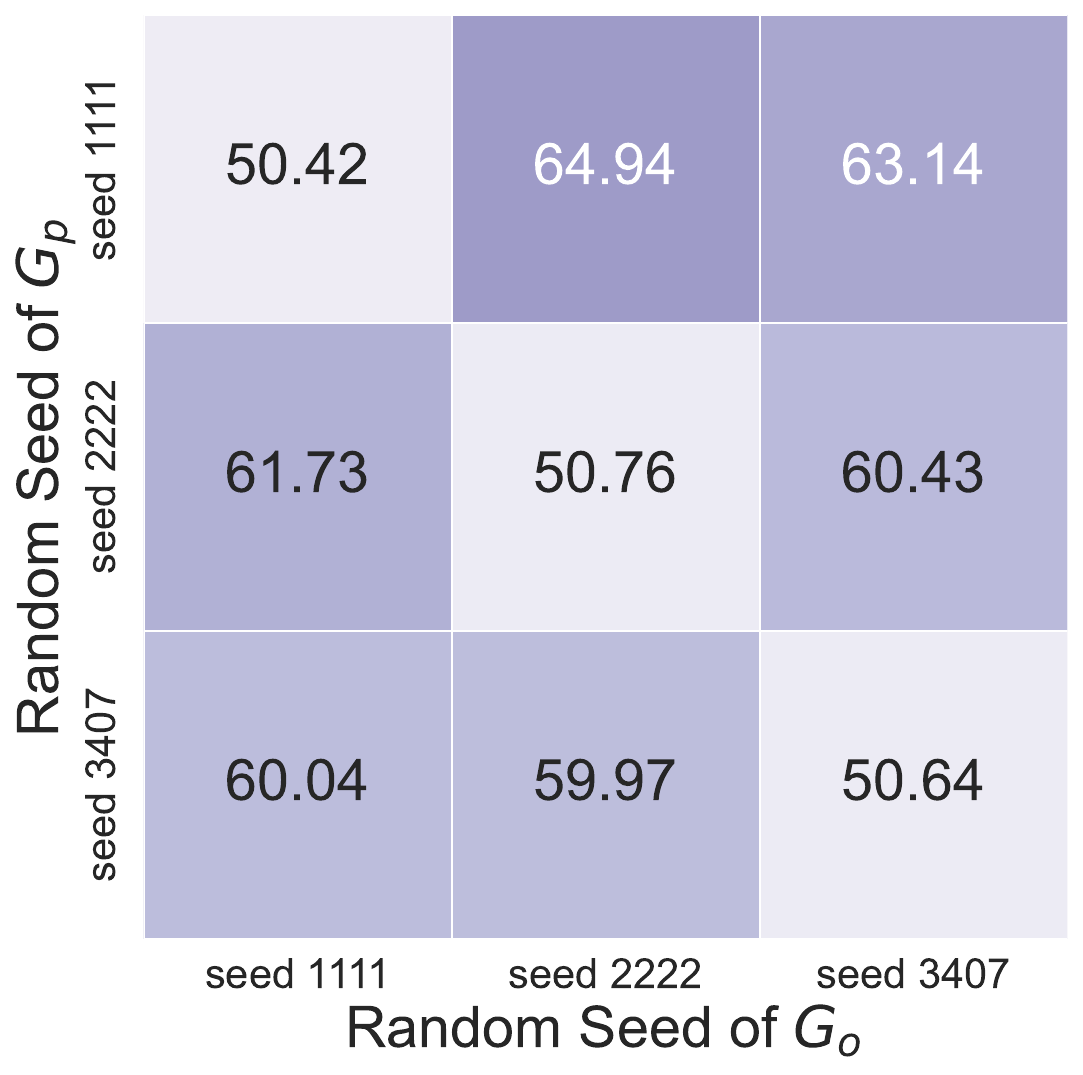}
         \includegraphics[height=0.45\linewidth]{img/cbar.pdf}
         \vspace{-5pt}
         }
         \subfigure[AUC scores on different LSUN (left) and CelebA (right) subsets.]{
         \centering
         \includegraphics[width=0.45\linewidth]{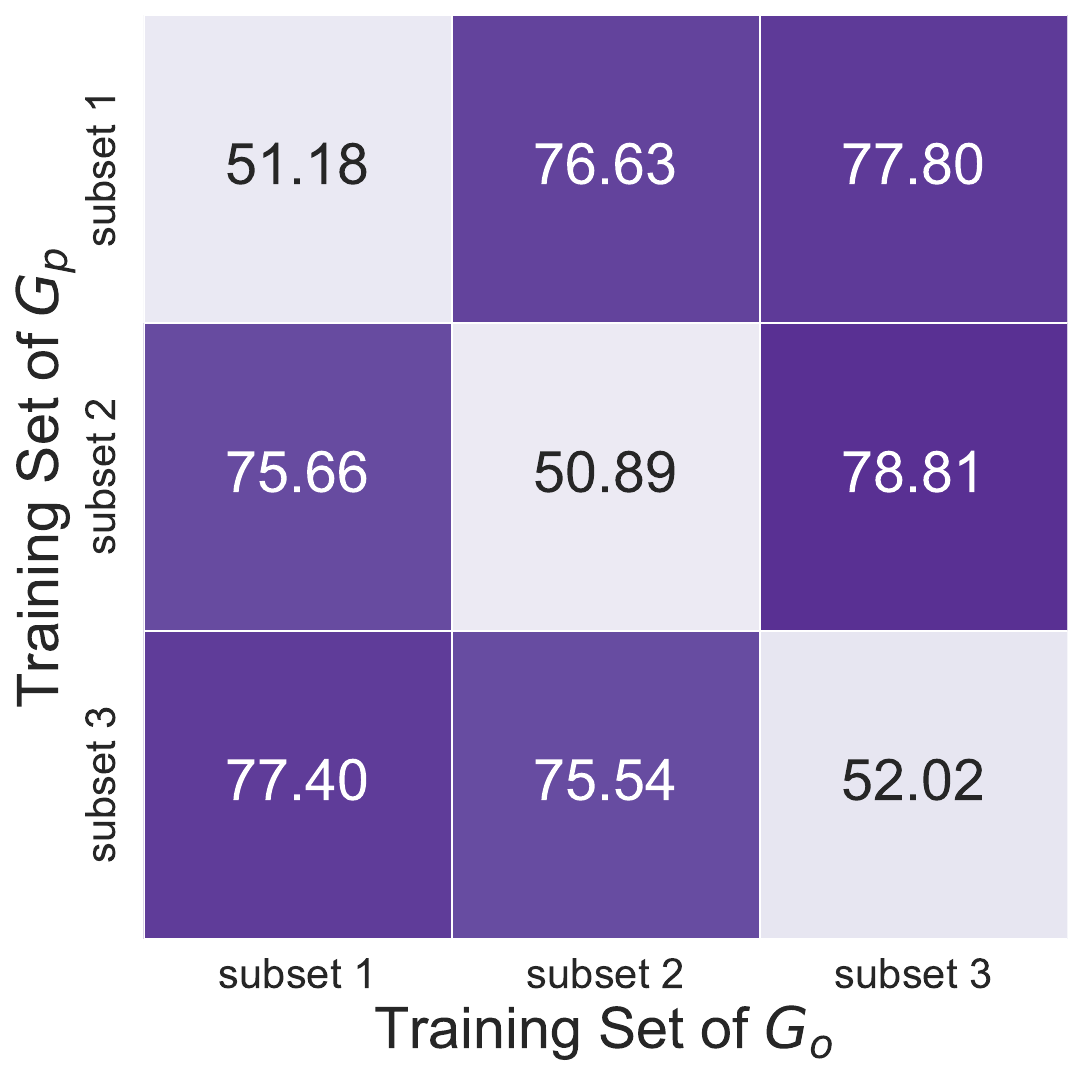}
         \includegraphics[width=0.45\linewidth]{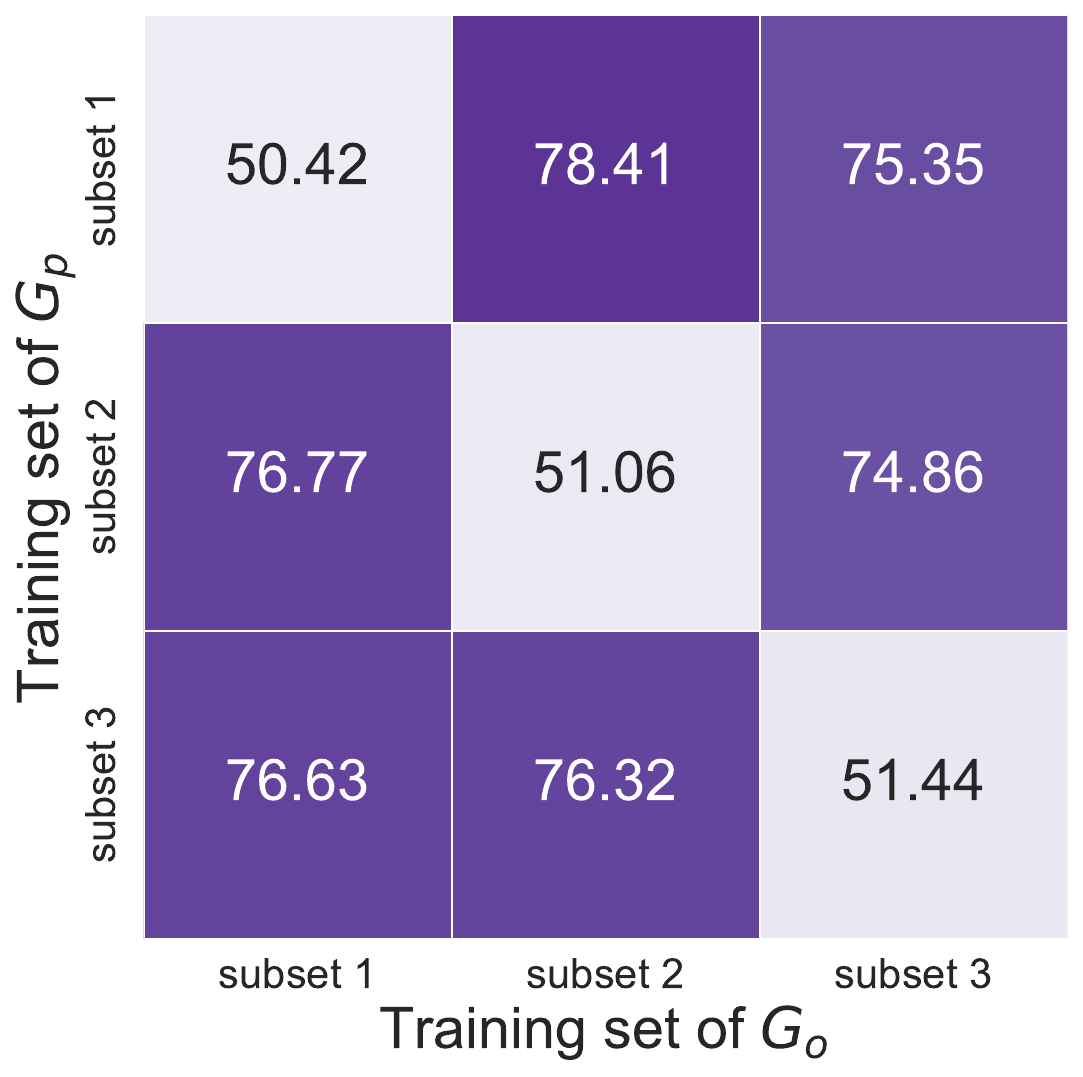}
         \includegraphics[height=0.45\linewidth]{img/cbar.pdf}
         }
     \caption{Effectiveness evaluation on image synthesis, in terms of three critical elements. The top panel indicates the evaluation on LSUN and the bottom panel denotes the evaluation on CelebA.  We use $D_{o}$ to distinguish source model $G_o$ and suspect model $G_p$. The results on the diagonal represent that the two models are identical (i.e. {$G_p$} is copied from $G_o$).}
     \vspace{-10pt}
     \label{fig:eff}
\vspace{-20pt}
\end{figure}

\section{Experiments}\label{sec:exp}
In this section, we mainly explore the effectiveness, scalability and robustness of our proposed approach. Some additional experiments and ablation studies of our method refer to \textcolor{magenta}{supplementary materials}. For the below experiments, we report the average result of ten independent replicates.
\subsection{Setup}

\noindent \textbf{Models.} For image synthesis, we use DCGAN \cite{radford2015unsupervised} and its two variants: SNDCGAN \cite{miyato2018spectral} and DCGAN established with residual block \cite{he2016deep}. We also conducted experiments on SOTA architectures including ProGAN \cite{karras2017progressive}, StyleGAN \cite{karras2019style}, StyleGAN2 \cite{karras2020analyzing} and StyleGAN3 \cite{karras2021alias}. For image-to-image translation, we consider three popular GANs that edit face attributes: StarGAN \cite{choi2018stargan}, AttGAN \cite{he2019attgan}, and STGAN \cite{liu2019stgan}. 

\noindent \textbf{Datasets.} We evaluate our method on two popular datasets, LSUN \cite{yu2015lsun} and CelebA \cite{liu2015faceattributes}. We use LSUN bedroom for image synthesis and CelebA for both image synthesis and image-to-image translation. 

\noindent \textbf{Evaluation Metrics.} For evaluating the effectiveness, we use the AUC score as mentioned earlier. We use structural similarity (SSIM) and Frechet Inception Distance (FID) \cite{borji2019pros} to measure image quality and similarity.

\subsection{Effectiveness}\label{subsec:effectiveness}
\begin{figure}[t]
    \centering
    \includegraphics[width=0.55\linewidth]{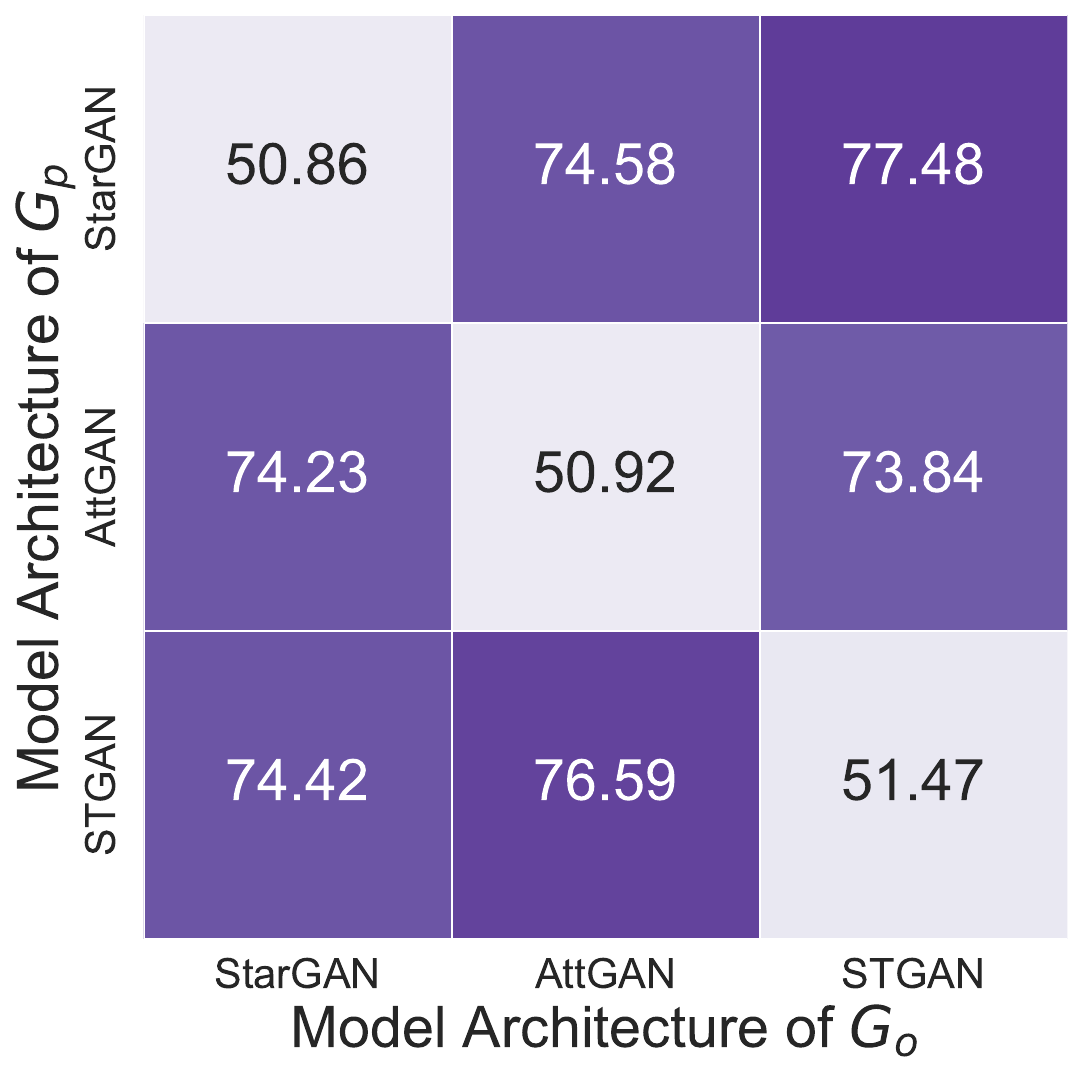}
    \vspace{-5pt}
    \caption{Effectiveness evaluation for image-to-image translation.}
    \label{fig:translation}
    \vspace{-10pt}
\end{figure}

We first show preliminarily that our method can effectively find out the piracy and would not mistakenly recognize homogeneous models (\ie{}, the model independently trained in similar settings) as piracy as well, even only small factors (\eg{}, initial seeds) are different. Fig. \ref{fig:eff} and Fig. \ref{fig:translation} present the experimental results on verifying the two popular GAN tasks (\ie{}, entire images synthesis, image-to-image translation) measured with AUC score. 

The closer the AUC score is to $100\%$, the greater the difference between two batches of synthesized outputs exposed. On the contrary, the closer the AUC score is to $50\%$, the more likely the two batches of images are from the same model because $\phi$ judges that there is no notable difference in the data distribution represented by the two batches.

\noindent \textbf{Model Architectures.} We consider the case where GANs with different architectures are trained on the same dataset and initial seeds. Experimental results in Fig. \ref{fig:eff} and \ref{fig:translation} illustrated that our method can identify the differences between GANs with different model architectures with an AUC score larger than $70\%$ and determine a piracy GAN with the same elements with an AUC score close to $50\%$.

\noindent \textbf{Training Datasets.} We also investigate the effect of training a GAN on different datasets. We respectively split CelebA and LSUN into 3 disjoint subsets with 50k images each. This ensures the training data are from the similar distribution. Fig. \ref{fig:eff} (c) shows that the AUC scores of two GANs trained on different datasets are all above $75\%$. The AUC score is higher than the other two elements, which indicates that the GAN training is sensitive to the training datasets.

\label{sec:exp2}
\noindent \textbf{Initialization seeds.} We finally investigate the effects of initial random seeds in ownership verification. Fig. \ref{fig:eff} shows that our method maintains AUC around $60\%$. Meanwhile, all the AUC scores of two identical models are less than $53\%$. We note that the initialization random seed is the smallest variable in training a homogeneous GAN. This extreme situation (\ie{}, the source model and the piracy only differ in random seeds) is almost impossible to happen in reality. However, our scheme still achieves an AUC score of $\sim 60\%$, notably margins from the AUC score of two same models ($\sim 50\%$). This is why we set the suspicious AUC score to $60\%$.

\begin{table}[h]
\scriptsize
\centering
\vspace{2pt}
\setlength{\tabcolsep}{3.5pt}
\begin{tabular}{c|c|c|c|c}
\toprule
 & StyleGAN2 & StyleGAN3 & StyleGAN & ProGAN  \\ \midrule
StyleGAN2 & \textbf{53.39} & 82.57 & 84.93 & 85.66\\
StyleGAN3 & 83.12 & \textbf{53.28} & 83.52 & 86.34
\\
\bottomrule
\end{tabular}
\vspace{-8pt}
\caption{Evaluation on unknown and SOTA GANs. The AUC means the classifier is trained with \{row\}'s discriminator, measuring the images generated by the \{column\}'s $G$. We mark the results from the paired $G$ and $D$ in \textbf{bold font} and unpaired in normal font.}
\label{tab:sota}
\vspace{-10pt}
\end{table}

\subsection{Scalability to SOTA Architectures}

Table \ref{tab:sota} shows the experimental results on the SOTA GAN Architectures. The results show that on SOTA architectures the performances are even better (AUC $> 80\%$). This is because the more complex the task is (\eg{}, larger datasets, higher resolution, more sophisticated network architecture), the discriminator has larger parameters and the distributions learned by different GAN instances are more complex thus more different from each other. This shows that our method generalizes well on SOTA and potential future architectures.

\subsection{Robustness}\label{subsec:robustness}

In this section, we mainly evaluate the robustness of our method in tackling the three existing attacks, \ie{}, model pruning, output transformations, and ambiguity attack. The experiments here are done on DCGAN trained on CelebA with size $128 \times 128$. Specifically, we note that the discriminator plays a similar role to private keys and should be kept secret by the owner. Therefore, according to our threat model, attacks that require the presence of the discriminator (\eg{}, fine-tuning) are not considered.

\begin{figure}[t]
    \centering
    \includegraphics[width=0.98\linewidth]{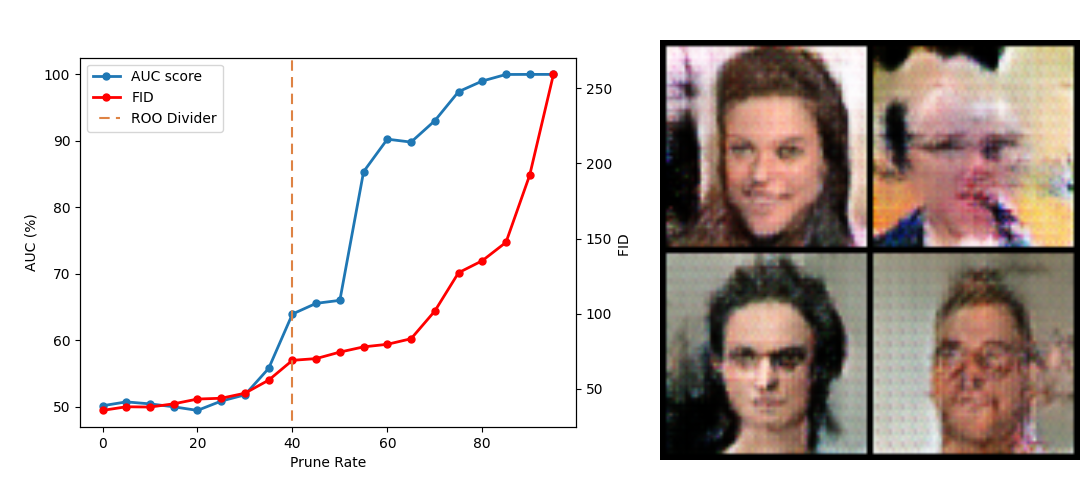}
    \vspace{-5pt}
    \caption{\textit{left}: Performance in evaluating the robustness against model pruning measured by AUC score. The dotted line is the ROO Divider, the operation region where acceptable image quality holds. \textit{right}: Visualization of synthesized images when the pruning rate is 40\%.}
    \label{fig:prune}
    \vspace{-10pt}
\end{figure}

\noindent \textbf{Pruning.} The model pruning aims to reset the the unimportant weights to 0 without introducing any performance degradation to the main task, like image synthesis in GANs. In experiments, we randomly reset the weights in GANs to explore whether our proposed method is sensitive to the model pruning. Experimental results in Fig. \ref{fig:prune} (left) shows the AUC score for measuring the similarity of two models is less than $60\%$ when the pruning rate reaches almost $40\%$. This indicates that our proposed method could verify two models in high confidence in this pruning rate settings. Here, the pruning rate less than $40\%$ is a \textit{region of operation} (ROO) as the quality of synthesized image is almost acceptable, where the FID score is larger than $69$. Fig. \ref{fig:prune} (right) visualizes the synthesized images when the pruning rate is $40\%$, which exhibits obvious damage. Thus, our method could survive the model pruning well.

\begin{table}[h]
\scriptsize
\centering
\setlength{\tabcolsep}{4.5pt}
\begin{tabular}{c||c|c|c|c}
\toprule
\multirow{2}{*}{\textbf{\makecell{Image transformation (magnitude)}}}
 & \multicolumn{2}{c|}{\textbf{CelebA}} & \multicolumn{2}{c}{\textbf{LSUN}} \\
& AUC & SSIM & AUC & SSIM \\ \midrule
noise ($\epsilon$=0.05) &  56.41 & 0.81 & 55.93 & 0.84\\
blur (ks=3, $\sigma$=2) &  57.16 & 0.78 & 57.24 & 0.80 \\
JPEG (factor=60)  &  54.90 & 0.73 & 56.77 & 0.69\\
crop (15\%) &  54.15 & 0.25 & 56.71 & 0.27 \\
\bottomrule
\end{tabular}
\vspace{-5pt}
\caption{Evaluation for the image transformation attacks measured by AUC scores and the similarity of images is measured by SSIM. The value of magnitude indicates the ROO, where the image has acceptable quality.}
\label{tab:transformation}
\vspace{-5pt}
\end{table}

\begin{figure*}[t]
\centering
\setlength{\abovecaptionskip}{0.2cm}   
\setlength{\belowcaptionskip}{-0.3cm} 
\subfigure[Original]{
\includegraphics[width=0.38\columnwidth]{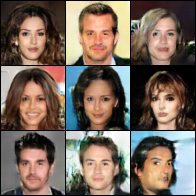}
}
\subfigure[Noise]{
\includegraphics[width=0.38\columnwidth]{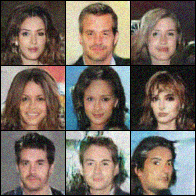}
}
\subfigure[Blur]{
\includegraphics[width=0.38\columnwidth]{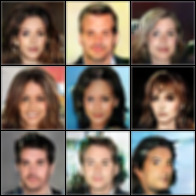}
}
\subfigure[JPEG]{
\includegraphics[width=0.38\columnwidth]{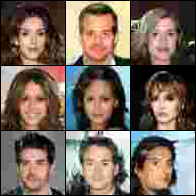}
}
\subfigure[Crop]{
\includegraphics[width=0.38\columnwidth]{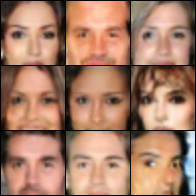}
}
\vspace{-5pt}
\caption{Visualization of synthesized images where the magnitude of image transformation is in the ROO (the value refer to Tab. \ref{tab:transformation}).}
\label{fig:transformations}
\end{figure*}

\begin{figure}[h]
\vspace{-10pt}
\centering
\setlength{\abovecaptionskip}{0.2cm}   
\setlength{\belowcaptionskip}{-0.3cm} 
\subfigure[Noise]{
\includegraphics[width=0.44\columnwidth]{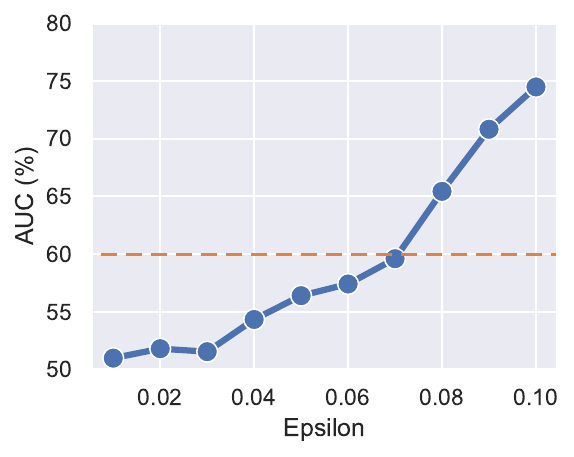}
}
\subfigure[Blur]{
\includegraphics[width=0.44\columnwidth]{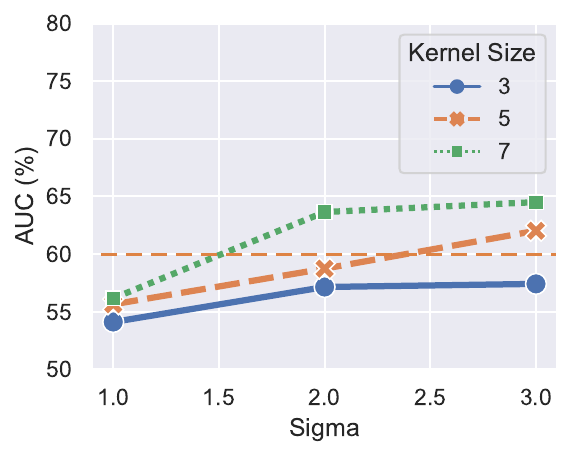}
}
\subfigure[JPEG]{
\includegraphics[width=0.44\columnwidth]{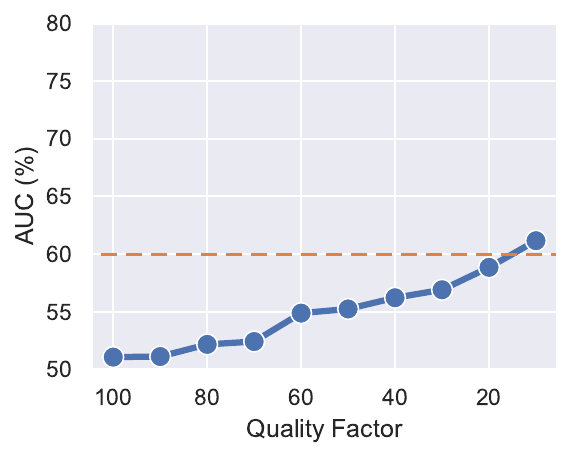}
}
\subfigure[Crop]{
\includegraphics[width=0.44\columnwidth]{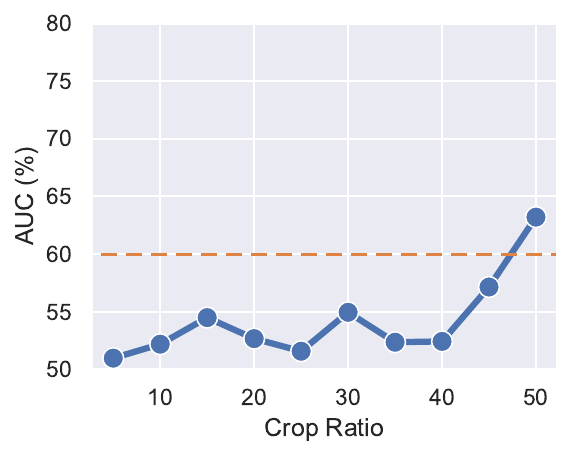}
}
\vspace{-5pt}
\caption{Four image transformation attacks under different intensities.}
\label{fig:transformations_curve}
\end{figure}

\noindent \textbf{Image Transformations.} Though input transformations are vain attempts, the adversary may conduct various output transformations to evade the verification. In experiments, we explore whether the sample transformation brings any degradation to our verification. Fig. \ref{fig:transformations_curve} shows the performance of our method measured by AUC score in identifying suspicious models under four types of image transformations. The dotted line in Fig. \ref{fig:transformations_curve} indicates a ROO where the synthesized images have no serious damage to human eyes. Fig. \ref{fig:transformations} visualizes the images under the ROO setting and Tab. \ref{tab:transformation} shows the corresponding magnitudes for the four types of image transformation. Experimental results illustrated that our method could identify the ownership effectively in the ROO of image transformations. We attribute this to the distribution-capturing property of our proposed method, which potentially learned some high-level features that are robust against these low-level transformations.

\noindent\textbf{Ambiguity Attacks.} In a real scenario, the adversary launches an ambiguity attack by obtaining a classifier that performs as well as the owner's one after replicating the model illegally. We simulate this attack by training a classifier using exactly the same technique described in Sec. \ref{sec:method} but without the help of the discriminator. Tab. \ref{tab:ambiguity attack} shows that random initialization will significantly degrade the performance of the learned hypersphere. The reason, as we have mentioned earlier, is that the discriminator provides a strong and unique center $\boldsymbol{c}$, compared to naive random initialization. Thus, the attacker failed to obtain a well-performed classifier even if he obtains the philosophy of our proposed method.

\begin{table}[h]
\scriptsize
\centering
\setlength{\tabcolsep}{3.5pt}
\begin{tabular}{c||c|c}
\toprule
Training strategy & AUC (same) $\downarrow$& AUC (different) $\uparrow$\\ \midrule
w/o The Discriminator (piracy) & 51.91 & 57.21 \\\midrule
w/ The Discriminator (Ours) & \textbf{50.18} & \textbf{75.63}\\
\bottomrule
\end{tabular}
\vspace{-5pt}
\caption{Performance in resisting ambiguity attacks. The column \textit{same} indicates the verification of two same GAN models, while the column \textit{different} denotes the verification for different GAN models.}
\label{tab:ambiguity attack}
\vspace{-10pt}
\end{table}

\vspace{-5pt}
\section{Conclusion and Discussion}
In this paper, we propose a novel ownership verification scheme for GANs, which working in box-free manner, universal to popular GAN tasks, and resisting the powerful ambiguity attack well. Inspired by the power of discriminator in witnessing the development of generators in synthesizing images gradually, we empower the discriminator to capture the unique GAN training which is important for ownership verification. Evaluation experiment results show that our method is highly effective, general and robust.

\noindent \textbf{Limitations and Discussions}. Though there seem to be no trivial adaptive attacks, our method relies on the empowered discriminator to capture the unique distribution learned by the generator. Therefore, if the discriminator is disclosed, the adversary may train a same classifier to confuse the verification, or leverage the discriminator to perform adversarial attacks. However, the discriminator is conventionally considered useless and will not be used after training. In the popular MLaaS scenario, there is also no reason for the model owner to open it to a third party. The adversary could not even access the discriminator through APIs. In our proposed scheme, the discriminator plays a similar role to private keys and should be kept secret by the owner. 

There are many spaces worth discovering for future works. For example, one may extend our insight of distribution capturing to  other generative models like diffusion models. It is also interesting to explore effective methods to evade our box-free  verification, which could be our future work. 
\vspace{-5pt}
\section{Acknowledgment}
This research was supported in part by the National Key Research and Development Program of China under No.2021YFB3100700, the National Natural Science Foundation of China (NSFC) under Grants No. 62202340, the Open Foundation of Henan Key Laboratory of Cyberspace Situation Awareness under No. HNTS2022004, Wuhan Knowledge Innovation Program under No. 2022010801020127, the Fundamental Research Funds for the Central Universities under No. 2042023kf0121, the Natural Science Foundation of Hubei Province under No. 2021CFB089.


\clearpage
\newpage

\balance
{\small
\bibliographystyle{plainnat}
\bibliography{ref}
}

\end{document}